\documentclass[conference]{IEEEtran}
\IEEEoverridecommandlockouts
\usepackage{cite}
\usepackage{amsmath,amssymb,amsfonts}
\usepackage{graphicx}
\usepackage{textcomp}
\usepackage{xcolor}
\def\BibTeX{{\rm B\kern-.05em{\sc i\kern-.025em b}\kern-.08em
    T\kern-.1667em\lower.7ex\hbox{E}\kern-.125emX}}

\usepackage{soul}
\usepackage{url}
\usepackage[hidelinks]{hyperref}
\usepackage[utf8]{inputenc}
\usepackage{multirow}

\usepackage{algorithm}
\usepackage{algorithmicx}
\usepackage[noend]{algpseudocode}
\usepackage[caption=false]{subfig}
\usepackage{booktabs}

\newtheorem{assumption}{Assumption}
\newtheorem{problem}{Problem}
\newtheorem{theorem}{Theorem}
\newtheorem{proof}{Proof}

\begin{document}

\title{Beyond Discriminant Patterns: On the Robustness of Decision Rule Ensembles
}

\author{\IEEEauthorblockN{1\textsuperscript{st} Xin Du}
\IEEEauthorblockA{\textit{School of Computing and Information Technology} \\
\textit{Great Bay University}\\
Dongguan, China \\
x.d.du@hotmail.com}
\and
\IEEEauthorblockN{2\textsuperscript{nd} Subramanian Ramamoorthy}
\IEEEauthorblockA{\textit{School of Informatics} \\
\textit{University of Edinburgh}\\
Edinburgh, UK \\
s.ramamoorthy@ed.ac.uk}
\and
\IEEEauthorblockN{3\textsuperscript{rd} Wouter Duivesteijn}
\IEEEauthorblockA{\textit{DAI Cluster} \\
\textit{Eindhoven University of Technology}\\
Eindhoven, Netherlands \\
w.duivesteijn@tue.nl}
\and
\IEEEauthorblockN{4\textsuperscript{th} Jin Tian}
\IEEEauthorblockA{\textit{Department of Computer Science} \\
\textit{Mohamed bin Zayed University of Artificial Intelligence}\\
Abu Dhabi, UAE \\
jin.tian@mbzuai.ac.ae}
\and
\IEEEauthorblockN{5\textsuperscript{th} Mykola Pechenizkiy}
\IEEEauthorblockA{\textit{Department of Mathematics and Computer Science} \\
\textit{Eindhoven University of Technology}\\
Eindhoven, Netherlands \\
m.pechenizkiy@tue.nl}
}

\maketitle

\begin{abstract}
Local decision rules are highly regarded for their interpretability, offering insights into granular patterns that are critical for explainable machine learning. While existing methods emphasize the identification of discriminative patterns to achieve high predictive accuracy, they often fail to account for robustness against distributional shifts that occur during deployment. This paper addresses this gap by proposing a novel approach to learning and ensembling local decision rules that are inherently robust across diverse training and deployment environments. Our method leverages causal inference principles, viewing distributional shifts as interventions on the underlying system. We incorporate two regularization techniques: graph-based regularization, which decomposes invariant features using causal graphs, and variance-based regularization, which promotes stability by introducing artificial features to guide decision boundaries. These techniques enable the generation of decision rules that excel in predictive power while maintaining stability under changing environmental conditions.
Extensive experiments on synthetic and benchmark datasets validate the effectiveness of the proposed method. The results demonstrate significant improvements in robustness, outperforming traditional boosting ensembles when subjected to diverse and challenging environments. Quantitative and qualitative analyses further highlight how the integration of causal knowledge and adaptive regularization encourages the utilization of invariant features, leading to better generalization. This work emphasizes the importance of causal reasoning in the design of machine learning models, paving the way for future research into robust, interpretable, and reliable decision-making frameworks for real-world applications.
\end{abstract}


\section{Introduction}

Local decision rule learning aims to discover the most discriminant patterns that can distinguish data points with different labels~\cite{morik2005local}. Local rules are commonly understood to be explainable and transparent~\cite{furnkranz2012foundations}. 
For example, in clinical health care, one can find rules like ``BMI $>$ 25 and Blood pressure $>$ 140 $\rightarrow$ Diagnostic Results''.
Domain experts can easily identify useful information to assist decision making.

Existing work mainly focuses on discovering either the most discriminant patterns~\cite{belfodil2018anytime} or fast and high-accuracy ensembles~\cite{chen2016xgboost}. 
Some recent research started to analyze the fairness of discovered subgroups~\cite{kalofolias2017efficiently,du2020fairness}, the causal rules that are reliable across different subgroups~\cite{budhathoki2021discovering}, and rules that can explain the prediction of a group of neurons~\cite{fischer2021s}. 
Some close work employs a small amount of data from randomized controlled trials to learn causal invariant random forests~\cite{zeng2021causal}, or verifies the robustness of tree ensembles with perturbation of samples~\cite{devos2021versatile}. 
The robustness of decision rule prediction on purely pooled data without knowing the environment label is still underexplored. We should investigate whether and how the local decision rules and their ensembles could perform robustly in deployment environments against the distributional shifts~\cite{kull2014patterns}.

We consider the problem of distributional shifts from a causal view, as the consequence of interventions on the underlying data generating systems. The 
goal is to learn a stable classifier that performs robustly across different environments. 
Existing work either focuses on mitigating the worst case performance of the model on potential environments~\cite{duchi2021learning}, 
which is called Distributional Robust Optimization (DRO); or focuses on learning causal invariant correlations that can be stable across different environments~\cite{gong2016domain,magliacane2017domain}. 
Solving the invariant learning problem without knowing any additional information is nearly impossible. Existing work requires to know environment labels or causal knowledge represented by Bayesian networks. We study this problem by assuming that the environment label is unknown, and the causal graph is partially known.
Our aim is to leverage  graph criteria that can guide the rule generating algorithms to construct highly predictive and robust decision rules.
Specifically, we focus on binary classification in this paper, though it is possible to generalize to multi-class classification.

\subsection{Main Contributions}
\begin{enumerate}
\item We  formulate the problem of robust local decision rule ensembles across environments. To learn robust decision rules, we propose two regularization terms to encourage the use of causal invariant features for rule generating. As far as we know, this is the first work that highlights the explainability and robustness of predictivity of decision rule ensembles.
\item We propose a regularization derived from a causal graph, and develop a soft mask feature selection method guiding the rule search algorithm. 
\item We propose a variance-based regularization by introducing artificial features and develop an iterative procedure to guide the rule generating algorithm.
\end{enumerate}
Experimental results on both synthetic and benchmark datasets demonstrate the effectiveness of our method quantitatively and qualitatively, showing that the two regularization terms are effective in encouraging the algorithms to use invariant features resulting higher robustness across different environments. Source code and datasets for reproducing the experiments are provided in \url{https://anonymous.4open.science/r/BDPRDRE-7785/}. 

\section{Problem Setup}
Assume a set of measurable features $X = \{X_1, \cdots, X_s\}$ and label $Y$. The joint $P(X,Y)$ is governed by an underlying system, where other unmeasurable variables $A$ exist. Following~\cite{arjovsky2019invariant}, we assume that
the datasets $\mathcal{D} = \{D_e\}_1^K$ are collected from a mixture of original environment and environments after interventions. Here $e \in \varepsilon$, where $\varepsilon$ denotes the environment space, and for each environment $e$, $D_e = \{x^i, y^i\}_1^{N^e}$. The robustness of a classifier $F$ will be measured as:
\begin{equation*}\label{eq:rbm}
\phi(F) = \max_{e \in \varepsilon} \mathbb{E}_{X,Y} [L(Y, F(X))|e].
\end{equation*}
Here, $L: Y \times \hat{Y} \rightarrow \mathbb{R}$ denotes a loss function that measures the performance of a model regarding $Y$, $\hat{Y}$, where $\hat{Y}$ is a random variable representing model predictions. Hence, the robustness of a model $F$ is represented by its worst case performance on the potential environments.

Further assume that the values of $X$ are taken from some domain $\mathcal{A}$.
A local decision rule can be defined as a function $r: \mathcal{A} \rightarrow \{-1, 1\}$. A data point is covered by rule $r$ as positive sample if and only if $r(x^i_1, \cdots, x^i_s) = 1$. The quality of a decision rule is measured by a function $\varphi(r) \rightarrow \mathbb{R}$, gauging discrimination of label $Y$. We mainly consider $Y$ as binary class, though it can be generalized to multi-class.
Now we formulate the main target problem for this paper:
\begin{problem}[Robust Decision Rule Ensembles]\label{pb:pb1}
Given a dataset $\mathcal{D} = \{D_e\}_1^K$, $D_e = \{x^i, y^i\}_1^{N^e}$ that comes from an environment governed by an underlying system and other environments after interventions. Without knowing labels of the environments, the task is to learn a classifier $F: \mathcal{X} \rightarrow \mathcal{Y}$ of decision rule ensembles, which can obtain highly predictive performance on the deployment environment that consists of any combination of the potential environments.
\end{problem}
The general way to solve this problem is to ensure the performance of the model on each potential environment by applying a measurement. However, since we cannot observe the environment label for the data at hand, we need additional assumptions about the prior knowledge or the structural relations as help. Following~\cite{gong2016domain}, we define the following assumptions:

\begin{assumption}[Invariant mechanism]\label{asp:asp1}
For a set of measurable features $X$ and target label $Y$ under a system with some unmeasurable variables $A$ which might affect both $X$ and $Y$, we assume that it is possible to decompose a subset $X^{\prime}$ from $X$ so that 
$\forall e \in \varepsilon, P(Y|X^{\prime}, e) = P(Y|X^{\prime}).$
\end{assumption}
Here we abuse $e$ to represent the potential interventions on the systems because we do not know where the interventions will be added.
\begin{assumption}[Sufficient mechanism]\label{asp:asp2}
$
\exists f, P(\hat{Y}|X^{\prime}) = f(X^{\prime}) + P(\epsilon),
$ 
so that
$\mathcal{L}(P(Y|X^{\prime}), P(\hat{Y}|X^{\prime})) < \delta$ holds.
\end{assumption}
Assumption~\ref{asp:asp1} ensures that it is possible to find a stable mechanism across different interventional environments. Assumption~\ref{asp:asp2} indicates that it is possible to learn a stable mapping function that is sufficient to predict the label $Y$ accurately using the invariant mechanisms.
Now we can formulate the robust decision rule ensembles as an optimization problem:
\begin{equation}\label{eq:obj}
\begin{split}
\min_{\{\alpha_1,\ldots,\alpha_M\}, \{r_1,\ldots,r_M\}} \max_{e \in \varepsilon} \mathbb{E}_{X,Y}&\left[L\left(Y,\sum_{m = 1}^{M}\alpha_m \cdot r_m(X)\right)\middle|e\right] \\
&+ ||\alpha||_2,\\
&\text{where}\ r_m = \text{arg}\max_r \varphi(r).
\end{split}
\end{equation}
The objective implies to find a set of rules and their associated coefficients that can minimize the maximal loss across all environments. Assumption~\ref{asp:asp1} encourages constructing rules with invariant features, so that the rules can have similar predictive performance across environments. Assumption~\ref{asp:asp2} pushes rules that are constructed with invariant features towards high predictive performance. 

\begin{figure}[t]
 \centering
   \subfloat[$\mathcal{G}$ \label{fig:genori}]{\includegraphics[width=0.25\columnwidth]{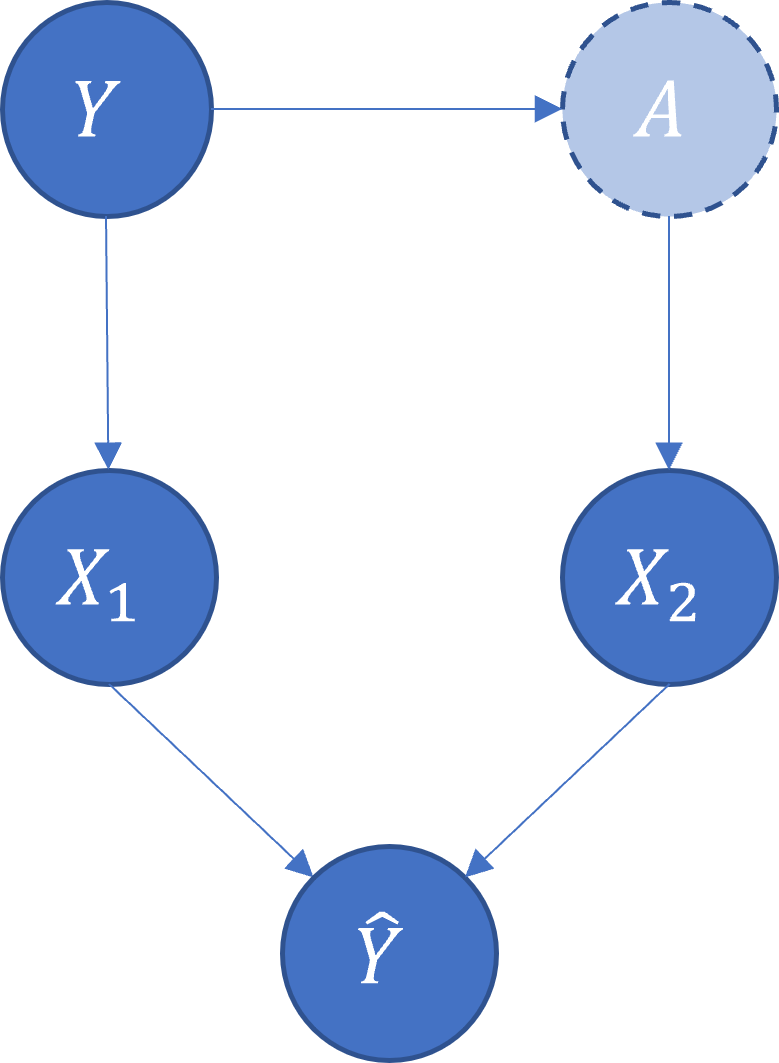}}
   \hspace{10mm}
   \subfloat[$\mathcal{G}^{\prime}$\label{fig:geninter}]{\includegraphics[width=0.25\columnwidth]{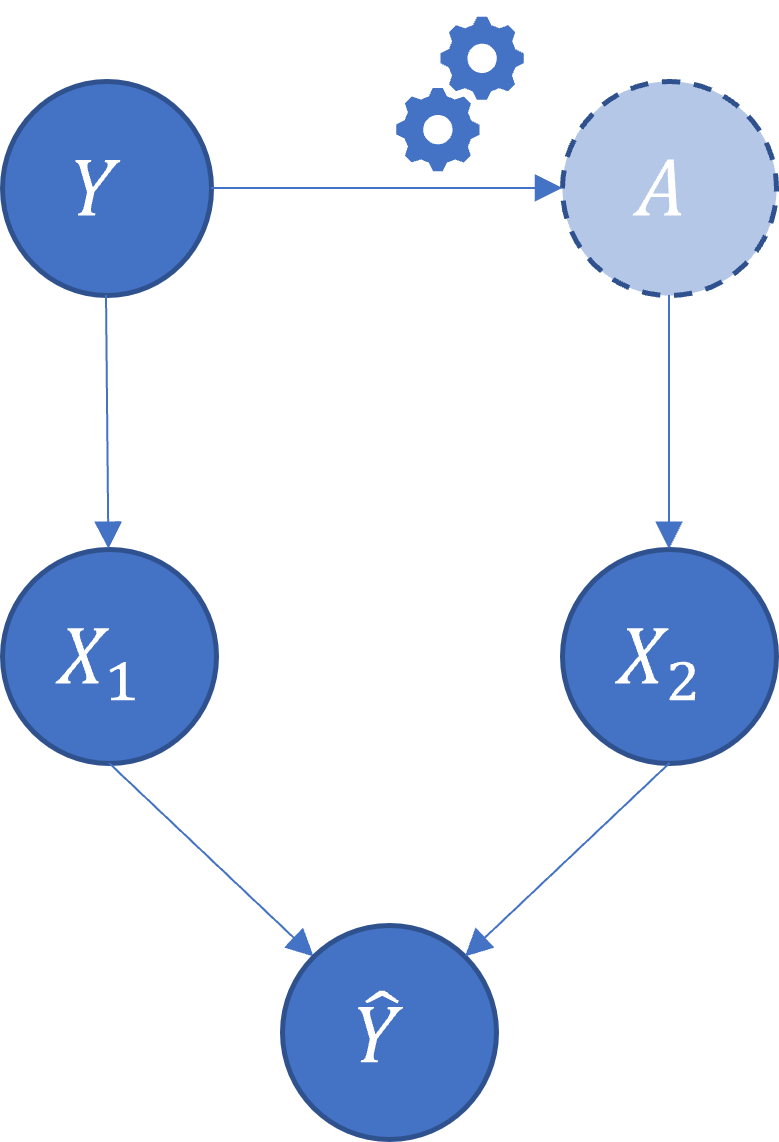}}
    \caption{Causal graphs representing the data generating process before and after intervention. In the right figure an intervention is conducted on the edge from $Y$ to $A$.}
    \label{fig:cg}
\end{figure} 

\subsection{Example 1}

\begin{figure}[!t]
 \centering
   \subfloat[Train data, non-robust classifier. \label{fig:exp1train}]{\includegraphics[width=0.385\columnwidth]{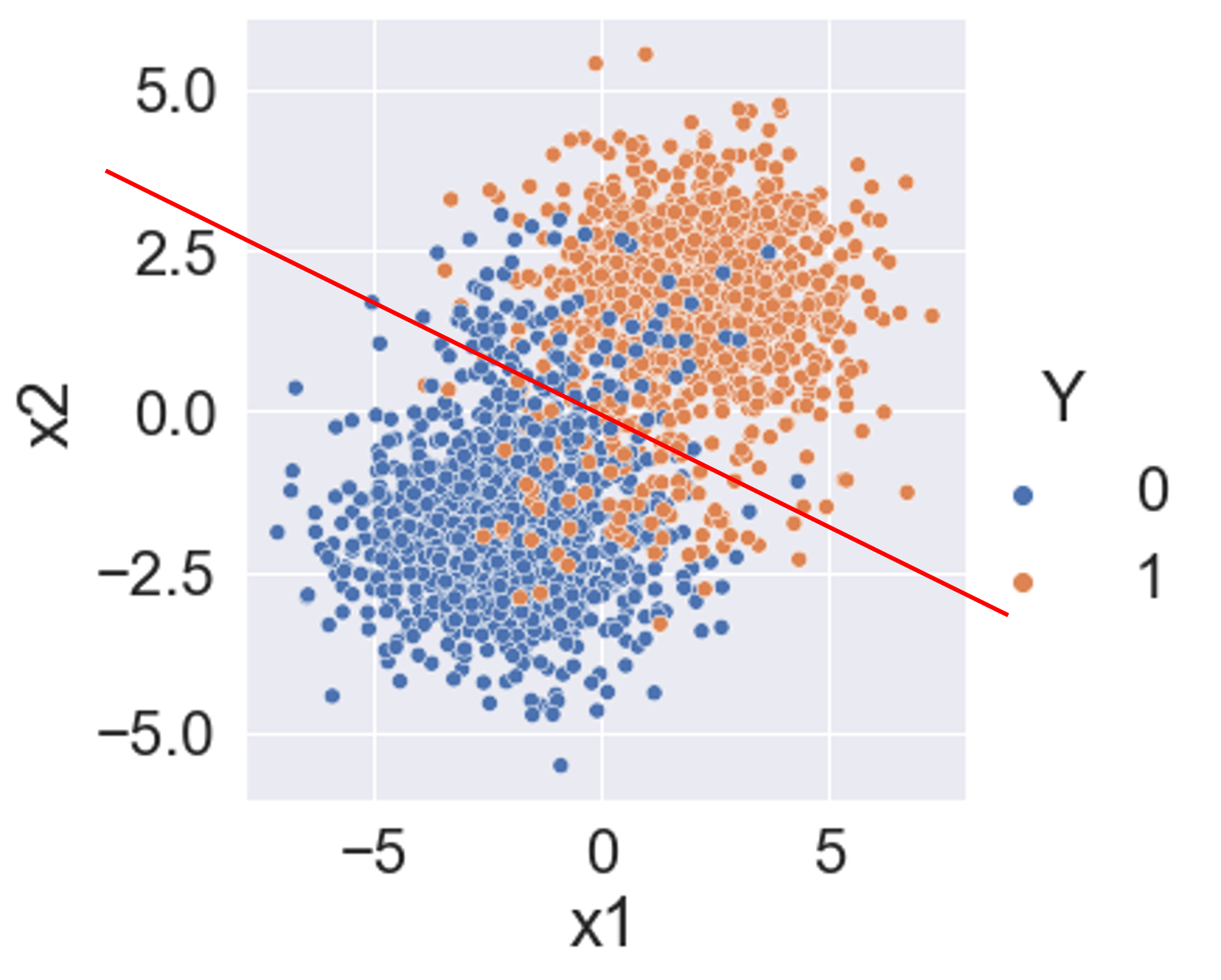}}
   \subfloat[Train and test data, robust classifier. \label{fig:exp1all}]{\includegraphics[width=0.415\columnwidth]{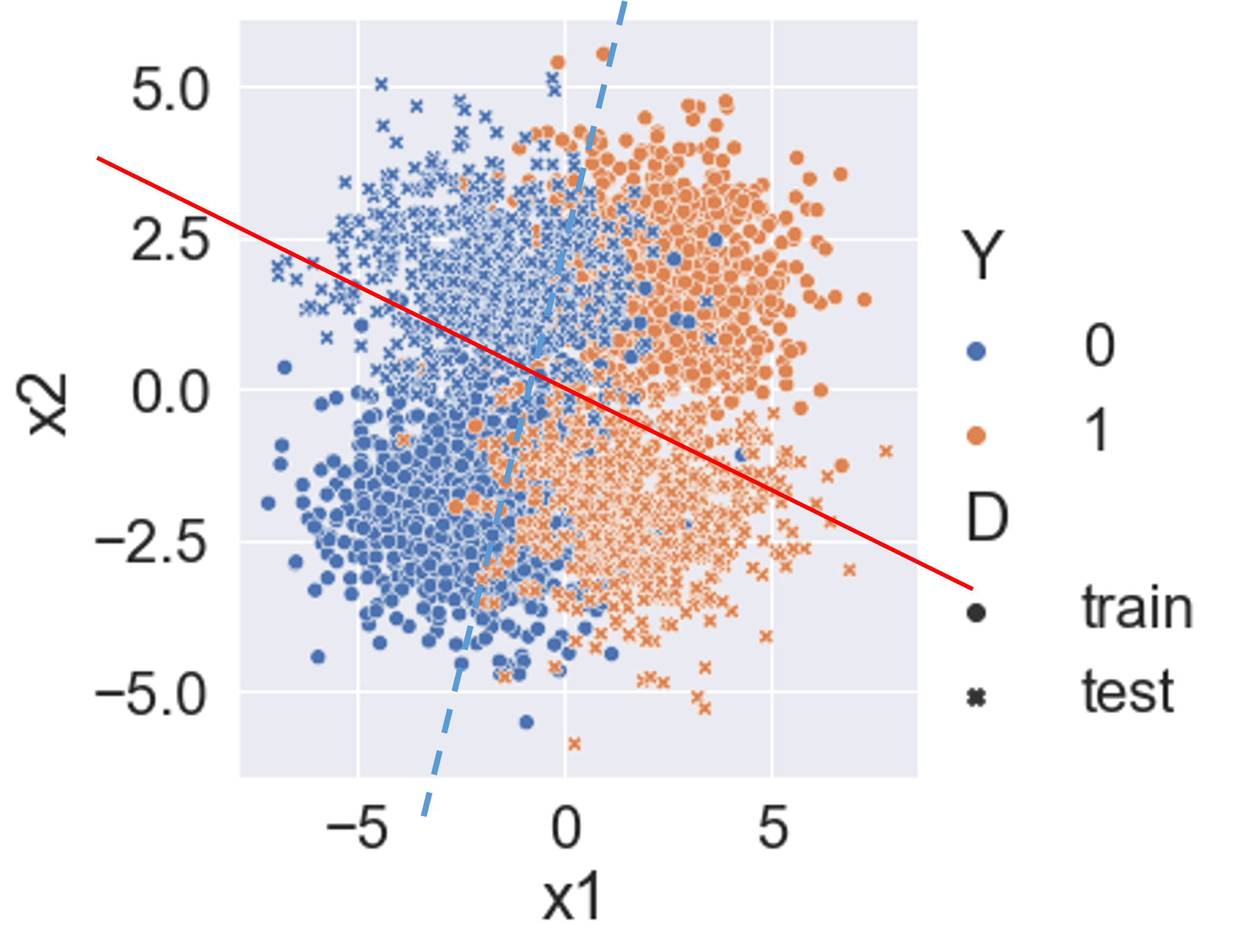}}
    \caption{Left: a classifier that overfits train data. Right: when adding test data from interventional environments, the previous classifier cannot distinguish target label anymore. A more robust classifier is shown as the dashed line. Data used in this case are generated by Example 1 with a 2-dimensional feature space.}
    \label{fig:exp1}
\end{figure} 

\begin{figure}[t]
 \centering
 \subfloat[Train.\label{fig:exp1res_train}]
   {\includegraphics[width=0.46\columnwidth]{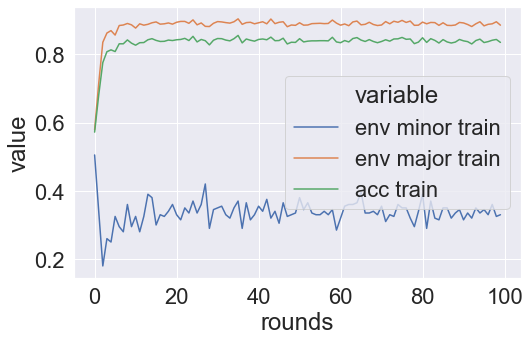}}
   \subfloat[Test.\label{fig:exp1res_test}]
   {\includegraphics[width=0.46\columnwidth]{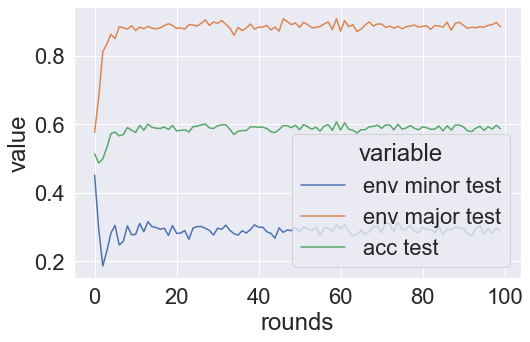}}
   \caption{Accuracy on train and test set, regarding to majority and minority environments. Data used in this case are generated by example 1 with a 10-dimensional feature space.}
    \label{fig:exp1res}
\end{figure}

We firstly give an example to show how traditional boosting rule ensembles fail on the problem.
Figure~\ref{fig:cg} illustrates distributional shifts modeled by an intervention
on variable $A$, which replaces the original function from $Y$ to $A$.
We assume as original data generating process:

\begin{equation*}
\begin{array}{ll}
\phantom{A|}Y \sim \text{Bernoulli}(U_0), & X_1|Y \sim \mathcal{N}(f_1(Y), \sigma_1),\\
A|Y \sim \mathcal{N} (f_2(Y), \sigma_2), & X_2|A \sim \mathcal{N}(f_3(A), \sigma_3),
\end{array}
\end{equation*}
where $U_0, \sigma_1, \sigma_2, \sigma_3$ are hyper-parameters. According to the local Markov condition~\cite{peters2017elements}, the joint distribution $P(X_1, X_2, Y, A)$ can be represented as:
\begin{equation*}
P(X_1, X_2, Y, A) = P(Y)P(X_1|Y)P(X_2|A)P(A|Y).
\end{equation*}
The intervention implies that $f_2(\cdot)$ is replaced with $f_2^{\prime}(\cdot)$, so that the joint distribution associated with $G^{\prime}$ (cf.\@ Figure \ref{fig:geninter}) is shifted by replacing $P(A|Y)$ with $P^{\prime}(A|Y)$. For the learning process, we assume that only $X = \{X_1, X_2\}, Y$ are observed; the task is to predict $Y$ with $X$.

We draw 90\% of train data from $P(X, Y)$ and 10\% from $P^{\prime}(X, Y)$; for test data we draw 50\% from both. The data are plotted in Figure~\ref{fig:exp1}. 
The rule ensemble algorithm we investigate is based on~\cite{friedman2008predictive}, and the rule search algorithm is from~\cite{duivesteijn2016exceptional}.
There are two iterative steps in the learning process: in the boosting step, local rule ensembles are used for loss computation; in the rule generating step, the beam search algorithm evaluates and generates high quality local rules. 

\begin{table}[t]
\centering
\caption{True positive rate of basis decision rules for the ensemble classifier.}
\label{tb:exp1rules}
\begin{tabular}{lcc}
\toprule
 $r$    & TPR Train   & TPR Test \\ 
\midrule
$x7 \leq -0.86$ and $x8\phantom{0} > -1.61$ & $0.83$     & $0.16$ \\
$x9 \leq -0.41$ and $x6\phantom{0} \leq -0.88$ & $0.85$     & $0.18$  \\
$x8 > \phantom{-}0.98$ and $x9\phantom{0} > \phantom{-}0.14$ & $0.73$     & $0.20$  \\
$x8 > -0.96$ and $x10 > -1.21$ & $0.81$     & $0.11$  \\
$x6 > -0.34$ and $x10 \leq -1.21$ & $0.79$     & $0.13$  \\\bottomrule
\end{tabular}
\end{table}

Given sufficient boosting iterations, accuracy on majority-environment data is high, but accuracy on data from the interventional environment is low (cf.\@ Figure~\ref{fig:exp1res}): the patterns in subgroups generated by the interventional environment are ignored, and the model doesn't generalize well in the presence of distributional shifts. Table~\ref{tb:exp1rules} lists 5 rules that are used to build the ensemble classifier: the TPRs of those rules reduce substantially on the test data.


 
\section{Methodology}

The core work of this paper is to investigate the robustness of decision rule ensembles under distributional shifts, and provide a method to ensure the robust performance across various environments. First of all, we need to look into details of the performance of the model.

\subsection{Robustness investigation}

\begin{algorithm}[t]
\caption{Local Decision Rule Learning.}
\label{al:alg1}
\begin{algorithmic}[1]
\Require Dataset $\mathcal{D}$, Quality Measure $\varphi$, Refinement Operator $\eta$, Integer $w$, $d$, $Q$, $M$, $N$,
\Ensure Classifier $F(X)$.
\Function{RuleGen}{$\mathcal{D}$, $\varphi$, $\eta$, $w$, $d$, $Q$, $Res$)}    
  \State candQ $\leftarrow$ new Queue;
  \State candQ.enqueue($\{\}$);            
  \State resultSet $\leftarrow$ new PriorityQueue($Q$);
  \While{level $\leq$ d}
    \State beam $\leftarrow$ new PriorityQueue($w$);
    \While{candQ $\neq$ $\varnothing$}
      \State seed $\leftarrow$ candQ.dequeue();
      \State set $\leftarrow$ $\eta$(seed);
      \ForAll{r $\in$ set}
        \State q $\leftarrow$ $\varphi(r)$; 
        \State resultSet.insert\_with\_priority(r, q);
        \State beam.insert\_with\_priority(r, q); 
      \EndFor
    \EndWhile
    \While{beam $\neq$ $\varnothing$}
      \State candQ.enqueue(beam.get\_from\_element());
    \EndWhile
  \EndWhile
  \State \Return{resultSet};
\EndFunction
\State
\Function{RuleEnsemble}{$\mathcal{D}$, $M$, $N$}
  \State $F_0(X) = 0$;
  \For{$m=1$ to $M$}
    \State $Res$ $\leftarrow$ $y_i - Sigmoid(F(x_i))$, $i = 1, \cdots, N$;
    \State $r_m, \alpha_m$ $\leftarrow$ RuleGen($\mathcal{D}$, $\varphi$, $\eta$, $w$, $d$, $Q$, $Res$) 
    \State $F_m(X) \leftarrow F_{m-1}(X) + \alpha_m \cdot r_m(X)$
    \State $F(X) \leftarrow F_m(X)$
  \EndFor
  \State \Return{$F(X)$};
\EndFunction
\end{algorithmic}
\end{algorithm}

From Equation~(\ref{eq:obj}), we can see that the classifier $F$ is a convex combination of local decision rules. 
The decision rule generating algorithm plays an important part in our method. In general, we apply a real-valued function $\varphi(\cdot)$ to evaluate each local decision rule. True positive rate and false positive rate are used to evaluate whether a rule can maximally discriminate the target label~\cite{furnkranz2005roc}. 

In Algorithm~\ref{al:alg1}, we implemented a quality measure based on that method. Let us recall Example 1; in Table~\ref{tb:exp1rules}, we list some of the rules that are used by the final classifier. As we can see, the true positive rate of those rules deteriorates in the data collected from environment after intervention. This is because the rule generating algorithm selected the variant features that are not against the intervention to construct the rule. 

So the question is: in what kind of situation could variant features be naturally selected rather than the invariant feature? The reason might be that the search algorithm is forced to maximize the discriminant patterns, so that the most informative features are selected. We have conducted multiple experiments by investigating the performance of the algorithm regarding to mutual information ratio between invariant and variant features. When the variant features are more informative than the invariant features, the models are very likely to fail on out-of-distribution (OOD) environments. 

\begin{figure}
 \centering
   \subfloat[1]{\includegraphics[width=0.46\columnwidth]{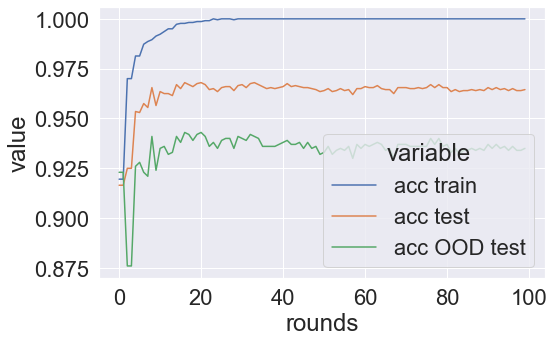}}
   \subfloat[3]{\includegraphics[width=0.43\columnwidth]{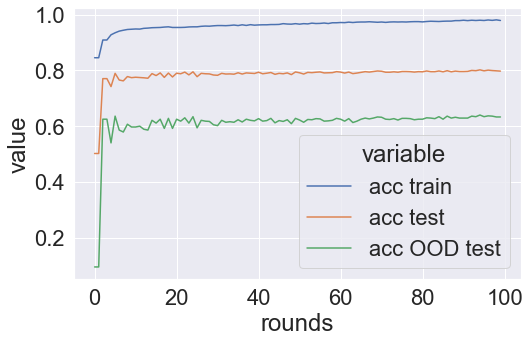}}
   \hspace{2mm}
   \subfloat[5]{\includegraphics[width=0.46\columnwidth]{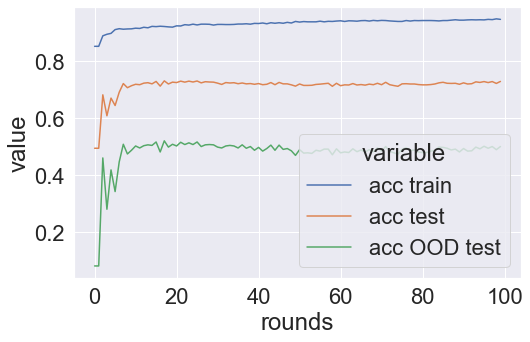}}
   \subfloat[7]{\includegraphics[width=0.46\columnwidth]{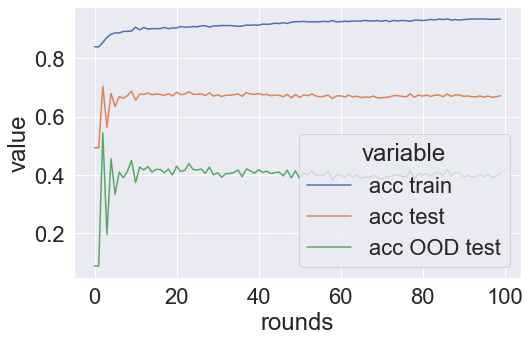}}
   \hspace{2mm}
   \subfloat[10]{\includegraphics[width=0.46\columnwidth]{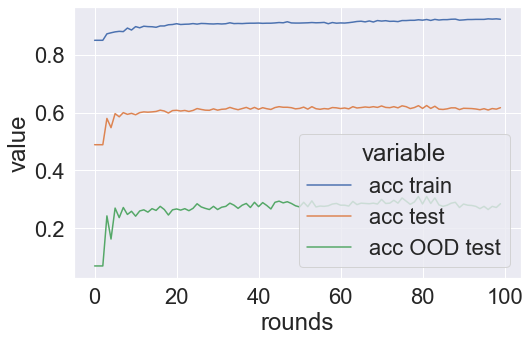}}
   \subfloat[15]{\includegraphics[width=0.46\columnwidth]{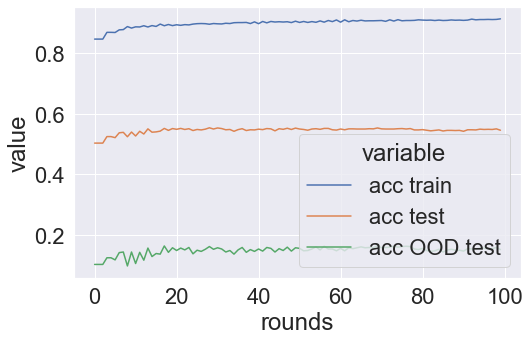}}
    \caption{Comparison for the performance of decision rule ensemble regarding to the mutual information ratio between variant and invariant features.}
    \label{fig:ratio}
\end{figure}

We apply a mutual information measure between features and the target label. Then according to the ratio of mutual information score between variant and invariant features, we run the model on different realizations of Example 1. The results are shown in Figure~\ref{fig:ratio}. As we can see, the performance of the model on OOD test environment are highly correlated with the ratio. This means that when the variant features are more informative than the invariant features in the given datasets, the models are very likely to fail on OOD environments. 

In order to mitigate this risk, we propose to apply regulation on the rule generating algorithm forcing it to learn the rules using causal invariant features. 

\subsection{Graph-based regularization}

The previous section shows that rule generating algorithms should be aware of the invariant mechanism while searching for the most discriminant features. Feature selection methods are usually known for their effectiveness on improving the generalization ability of the model on OOD data~\cite{chandrashekar2014survey}. We propose a graph-based method to regularize the feature selection process in the rule generating algorithm. 
The core idea is to decompose the joint distribution into a product of conditional distributions of each c-component~\cite{correa2019statistical,pearl2009causality,tian2002testable}.
Each variable in a c-component is only dependent on its non-descendant
variables in the c-component and the effective parents of
its non-descendant variables in the c-component.

\begin{theorem}\label{tm:cid}
Let $X, Y \subseteq V$ be disjoint sets of variables. Let $W = An(X \cup Y)_{\mathcal{G}}$ be partitioned into c-components $C(W) = \{C_1(W_1),\ldots,C_H(W_H)\}$ in causal graph $\mathcal{G}_{[An(X \cup Y)]}$.
Let $A$ be a set of manipulable variables for potential interventions. Then $X^{\prime} \in C_h(W_h)$ is an invariant feature across interventional environments when either 
$C_h(W_h) \cap A = \emptyset$, 
or
$C_h(W_h) \cap De(Y) = \emptyset,$ 
where $De(Y)$ represents the descendants of $Y$.
\end{theorem}
We do not specify the direction between $X$ and $Y$. Hence, if there is no direct intervention on $Y$, then features that are parents of $Y$ can be identified as invariant features. Another implication is that if there is an intervention on the Markov blanket of $Y$ that blocks the path from $Y$ to the feature, then we should consider the feature as a variant feature. 

Given a causal graph by domain knowledge, we can separate invariant subset of features $X^{\prime}$ from $X$. Then the general method is to build a mask set $\{\beta_1,\ldots,\beta_s\}$ where $\beta \in \{0, 1\}$ for the feature selection. The optimization problem in Equation~(\ref{eq:obj}) can be reformed as:
\begin{equation}\label{eq:obj2}
\begin{split}
\min_{\{\alpha_1,\ldots,\alpha_M\}, \{r_1,\ldots,r_M\}} \max_{e \in \varepsilon} \mathbb{E}_{X,Y}&\left[L\left(Y,\sum_{m = 1}^{M}\alpha_m \cdot r_m(X)\right)\middle|e\right] \\
&+ ||\alpha||_2 + ||\beta||_0.\\
\end{split}
\end{equation}
In the rule construction process, we construct the candidate rule by enumerating features one by one. By applying feature masks, the variant features are filtered in this step. Finally, the high quality features are kept for the next beam search iteration. 
Such a hard selection method might cause high variance and information loss~\cite{maddison2016concrete}. Following~\cite{yamada2020feature}, we adapt the soft feature selection method for rule generating algorithms, using Gaussian-based continuous relaxation for Bernoulli variables $\beta$. Here, $\beta_s$ is defined by $\beta_s = \max(0, \min(1, \mu_s + \sigma_s))$, where $\sigma_s \sim \mathcal{N}(0, \sigma^2)$. 
For $X_s \in X^{\prime}$, we set $\beta_s = 0$. For $X_s \in X \setminus X^{\prime}$, we set $\mu_s$ empirically, according to the mutual information ratio between $I(X_s;Y)$ and $I(X_{inv};Y)$, where $X_{inv}$ represents $X \setminus X^{\prime}$. Then we sample a $\mu_s$ to compute $\beta_s$.

In rule generating algorithm we apply Laplace or M-estimate~\cite{cestnik1990estimating} to measure the quality of discriminant patterns.  We define the following quality measure:
\begin{equation}\label{eq:qmbeta}
\varphi(r) = \frac{p + 1}{p + n + 2} - \lambda \cdot \beta_s,
\end{equation}
where $p$, $n$ denote the positive and negative samples that are covered by the rule, and $\lambda$ is scale parameter. By replacing $\varphi(r)$ in the original algorithm with Equation~(\ref{eq:qmbeta}), we encourage the algorithm to use rules that can maximize the discriminant patterns and consider the variant penalties.

To explain the search process based on Equation~(\ref{eq:qmbeta}), recall Example 1: in the search process, we treat numeric descriptors with the \texttt{lbca} discretization method from~\cite{meeng2021for}, using 8 equal-width bins.  
Suppose, e.g., we have base rule $r =$ `$x_2 > 0$  and $x_2 < 1$', then we compute the quality of this rule using Equation~(\ref{eq:qmbeta}). The highest quality rules are kept in the beam for the next search level. In the next iteration, we enumerate all the features again. Because the use of the regularization term, in Example 1, the quality of rules including $X_2$ are penalized, so that the algorithm is keen to construct rules with $X_1$. Then the top rule returned by the beam search algorithm is passed to `RuleEnsemble()' in each boosting step. Please refer to the Appendix~\ref{app:ifd} for more details of our algorithms.

\subsection{Variance-based regularization}

\begin{figure}[t]
 \centering
  \includegraphics[width=0.5\columnwidth]{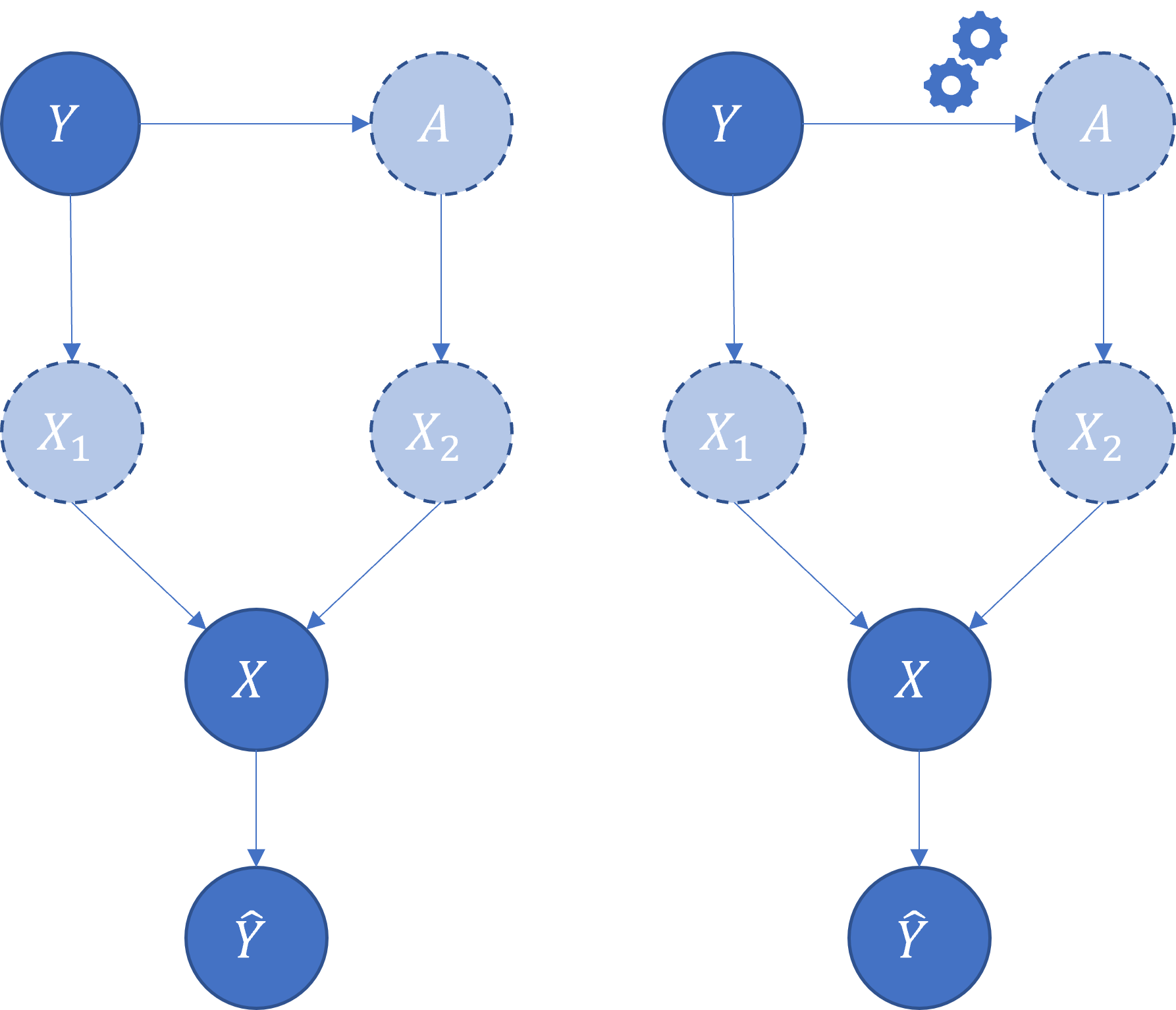}
  \caption{Example causal graph with unobserved features.}
    \label{fig:exp2}
\end{figure} 

In realistic scenarios, it is usually difficult to obtain accurate causal graphs.
In some situations, we cannot observe the features that contribute to the mixed features. Figure \ref{fig:exp2} shows an example where $X_1$ and $X_2$ are unobservable, while $X$ is derived from the combination of $X_1, X_2$. It is impossible to decompose invariant features from $X$ directly from the graph. As a consequence, we cannot build masks for feature selection. This usually happens on image datasets. Invariant features include the structures of the objects and variant features include color, background, or lightness.

Previous research~\cite{janzing2019causal} 
points out that, by applying regularization properly, a general regressor can gain the causal invariant property. Following this route, we aim to set up a variance-based regularization to build a classifier that uses causal invariant mechanisms. Specifically, we would like to artificially construct a new feature, which is stable with regard to environments and target labels. In Figure \ref{fig:exp2}, we can assume that the new feature is the child of $Y$ and parents of $X_1, X_2$. This feature can be used as a proxy to regularize the learning algorithm to employ the causal invariant features for prediction. The variance-based regularization term is adapted according to~\cite{heinze2021conditional}:
\begin{equation*}
\begin{split}
C_{F, \nu} &= E[\text{Var}(F(X)|Y, ID)^{\nu}], \\
C_{L, \nu} &= E[\text{Var}(L(Y,F(X))|Y, ID)^{\nu}],
\end{split}
\end{equation*}
where ID refers to the identifier variable that is the artificially constructed new feature variable. Now, the optimization problem can be reformed as:
\begin{equation}\label{eq:obj3}
\begin{split}
\min_{\{\alpha_1,\ldots,\alpha_M\}, \{r_1,\ldots,r_M\}} \max_{e \in \varepsilon} \mathbb{E}_{X,Y}&\left[L\left(Y, \sum_{m = 1}^{M}\alpha_m \cdot r_m(X)\right)\middle|e\right] \\
&+ ||\alpha||_2 + C,\\
\end{split}
\end{equation}
where $C$ is a general representation of $C_{F, \nu}$, $C_{L, \nu}$. Here the challenges are two-fold. On the one hand, as opposed to neural networks (where the regularization term is used to guide the optimization of weights), in our case, the regularization should be used for the rule generating algorithm.
On the other hand, the variance is computed on global data, while the traditional rule search algorithm only considers local discriminant patterns. How to enclose the variance to guide the search of local rules towards further benefiting the ensembles, is a challenge.

Specifically, we can compute $C_{L,1}$ as:
\begin{equation*}
\begin{split}
\hat{C}_{L,1} = &\frac{1}{q}\sum_{j=1}^q \frac{1}{|G_j|} \sum_{i\in G_j}(L(Y, F(x_i)) - \hat{\mu}_{L,j})^2,\\
&\text{where}\ \hat{\mu}_{L,j} = \frac{1}{|G_j|} \sum_{i \in G_j} L(Y, F(x_i)).
\end{split}
\end{equation*}
We employ the exponential loss function,
where for each new candidate rule in step $m$, we have:
\begin{equation}\label{eq:eloss}
e^{-Y \cdot F_m(X)} = e^{-Y \cdot (F_{m-1}(X) + \alpha_m r_m(X))}.
\end{equation}
By this definition, we can iteratively compute the variance term in each boosting loop for each candidate rule. Then the quality measure can be reformulated as:
\begin{equation}\label{eq:qmvar}
\varphi(r) = \frac{p + 1}{p + n + 2} - \lambda \cdot C_{L,F_{m-1},r_m,\nu}.
\end{equation}
This enables the search algorithm to select local decision rules with highly discriminant patterns as well as low global variance in each boosting step. Rather then only considering TPR and FPR, with Equation~(\ref{eq:qmvar}), we can compute the influence of the new added rule to the global level by computing the variance. Equation~(\ref{eq:eloss}) allows us to reuse the results in previous boosting steps, reducing the computational cost.
The same method can be applied to compute $C_{F, \nu}$. Please refer to the Appendix~\ref{app:vbr} for more details of our algorithms.



\section{Experiments}
In this section, we design multiple experiments to validate our method against the following research  questions:

\textbf{RQ1} How effectively can graph-based regularization ensure the robustness of decision rule ensembles?

\textbf{RQ2} How effectively can variance-based regularization ensure the robustness of decision rule ensembles?

\textbf{RQ3} What kind of rules can our algorithms retrieve that generalize well across various environments?

\subsection{Datasets}


\paragraph{`small invariant margin'} This benchmark is from~\cite{aubin2021linear}, as a linear version of the spiral binary classification problem originally introduced by~\cite{parascandolo2020learning}. The basic idea is to create variant features with large-margin decision boundary and invariant features with small-margin decision boundary. For each specific environment $e \in \varepsilon$, the data generating process is: $y \sim \text{Bernoulli}(U_0)$, $\text{when} \quad y = 0$, $x_{inv} \sim \mathcal{N}(\gamma, 10^{-1}), \quad x_{spu} \sim \mathcal{N}(\mu, 10^{-1})$, $\text{when} \quad y = 1$, $x_{inv} \sim \mathcal{N}(-\gamma, 10^{-1}), \quad x_{spu} \sim \mathcal{N}(-\mu, 10^{-1})$, where $\mu \sim \mathcal{N}(0, 1)$, $\gamma = 0.1 \cdot 1_{d_{inv}}$.


For different parts, we sample a different $\mu$, so that the patterns in variant features vary across environments.
The input features can be scrambled by a fixed random rotation matrix across potential environments. This prevents the algorithms from using a small subset to make invariant predictions. We generate datasets both with and without scrambles.

\paragraph{`cows versus camels'} This benchmark is also from~\cite{aubin2021linear}, as an imitation version of binary classification problem for cows and camels. The variant feature is denoted by the background~\cite{beery2018recognition}. The intuition is `most cows appear in grass and most camels appear in sand'. 


For benchmarks `cows versus camels' and `small invariant margin', we sample the datasets using open source from Github \url{https://github.com/facebookresearch/InvarianceUnitTests}.

\paragraph{Color MNIST} For this benchmark, we following the original version from~\cite{arjovsky2019invariant}. The data generating process is: $\text{if digit} (Y) < 5: Y_{obs} = 0$, $\text{else}: Y_{obs} = 1$, $t \sim \mathcal{U}(0,1)$, $\text{If} \quad t < 0.25: Y_{obs} = -Y_{obs}$, $X_{fig} \sim f_1(Y) + \sigma$, $\text{color C}(X_c) = f_2(Y_{obs}) + f_3(A)$,
$X_{obs} = X_{fig} \times [X_c,(1-X_c),0]$, $X_{obs}^{\prime} = \text{SVD}(X_{obs})$, where SVD represents Singular value decomposition~\cite{wall2003singular}. 


\begin{figure}[t]
 \centering
  \includegraphics[width=0.46\columnwidth]{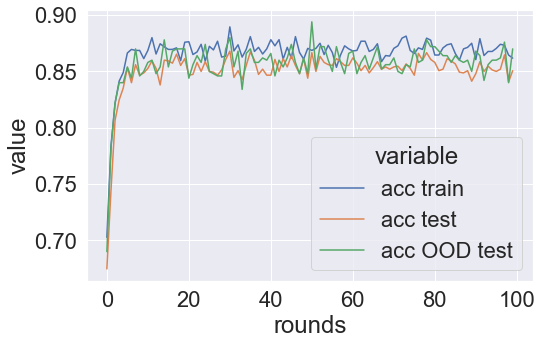}
  \caption{Accuracy on train, test, OOD test for synthetic data in Example 1 with graph-based regularization.}
    \label{fig:exp1res_graph}
\end{figure} 

\begin{table*}[t]
\centering
\caption{Example 1, TPR and FPR of basis decision rules for the ensemble classifier.}
\label{tb:exp1_graphrules}
\begin{tabular}{lcc}\toprule
 $r$    & Train TPR, FPR   & Test TPR, FPR  \\ 
\midrule
$x8 > 1.01$ and $x3 > -1.90$ or $x2 > 3.06$ & $0.56$, $0.08$    & $0.50$, $0.18$ \\
$x3 > -2.23$ and $x4 > -1.88$ and $x5 > -3.57$  & $0.80$, $0.15$ & $0.71$, $0.22$ \\
$x1 > -2.69$ and $x2 > 0.71$ or $x1 > 3.96$ & $0.71$, $0.14$   & $0.69$, $0.23$ \\
$x9 > -0.66$ and $x3 > 0.31$ or $x1 > 2.22$     & $0.70$, $0.19$ & $0.59$, $0.26$ \\
$x4 > 1.30$ or $x5 > 2.72$ or $x2 > 5.59$  & $0.80$, $0.19$  & $0.68$, $0.21$\\\bottomrule
\end{tabular} 
\end{table*}

\paragraph{Real-world Datasets} Apart from these semi-synthetic benchmarks, we also include several real-world tabular datasets with natural domain shifts~\cite{nastl2024causal}: \textbf{Adult Income}~\cite{adult_2}, this is a prediction task for predicting annual income. We choose the Race level as the domain shift identifier. \textbf{Bank Marketing}~\cite{bank_marketing_222}, this data is related with direct marketing campaigns (phone calls) of a Portuguese banking institution. We choose the Marital Status as the domain shift identifier. \textbf{Churn Modeling}, this data is from Kaggle.com, the target is to predict whether a customer will exit from the program. The shift identifier that we choose is Geography level, where we set Geography not equal to Spain as the majority group. \textbf{Credit Card}~\cite{default_of_credit_card_clients_350}, this dataset aims at the case of customers' default payments in Taiwan. We choose Education level as the domain shift identifier. \textbf{MimicIII Mort}~\cite{johnson2016mimic}, this data is from a freely-available database comprising deidentified health-related data associated with over forty thousand patients who stayed in critical care units of the Beth Israel Deaconess Medical Center between 2001 and 2012. We choose the insurance type as the domain shift identifier. \textbf{Readmission}~\cite{gardner2024benchmarking} this data is from clinical care at 130 US medical facilities, including hospitals and other networks. The goal is to predict whether a diabetic patient is readmitted to the hospital within 30 days of their initial release. We choose the admission source as the domain shift identifier.



\subsection{Implementation details}
We mainly consider beam search parameter: width $w$, depth $d$ and regularization scale $\lambda$; the boosting rounds $M$; and the environmental parameter: number of environments $K$. We fix the beam search width to $30$, depth to $2$ and boosting rounds to $100$. Experiments are varied by different regularization scale $\lambda$ and number of environments $K$. 
For all the experiments, we collect training data from one majority environment and several minority environments. For test data, we use data from several potential environments. 
We run the experiments on multiple synthetic and benchmark datasets. All the experiments are conducted on a single CPU (i5-11600K) desktop.

\subsection{Experiments on graph-based regularization}

To validate \textbf{RQ1}, we design experiments on synthetic data based on Example 1, and a public benchmark. 


We would like to test the performance of our method on multiple environments. Hence, we first sample a set of $K$ environments $\{e\}^K$. Then, according to the data generating process $P(X,Y|e)$, we sample data for each of the environments $\{D_e\}_1^K$, $D_e = \{X^i, Y^i\}^{n_e}_1$. Similar as Example 1, in each generating loop, we sample invariant feature set $X_{inv}$ and variant feature set $X_{var}$ then concatenate them to construct the observable features. Data sampled from various environments are scrambled by a full rank matrix. By this setting, the pooled data prevents us from using a small subset to construct the invariant predictor.

\begin{figure}[t]
 \centering
   \subfloat[before regularization.\label{fig:exp3_e3_naive}]
   {\includegraphics[width=0.46\columnwidth]{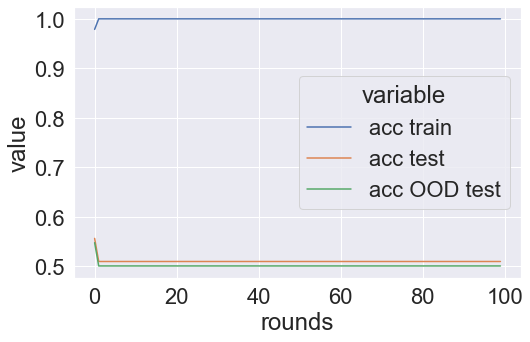}}
   \subfloat[graph-based regularization.\label{fig:exp3_e3_graph}]
   {\includegraphics[width=0.46\columnwidth]{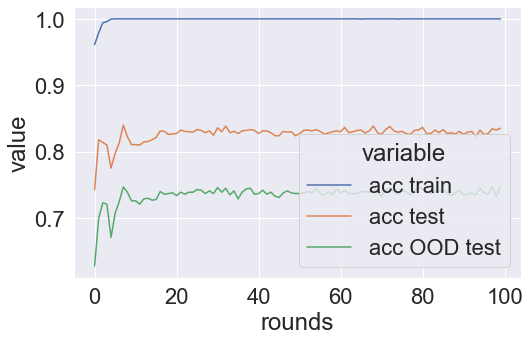}}
   \caption{Accuracy on train, test, OOD test for `small invariant margin' benchmark dataset with 3 environments.}
    \label{fig:exp3res_e3_graph}
\end{figure}

\begin{table*}[t]
\centering
\caption{Public benchmark (`small invariant margin'), TPR and FPR of basis decision rules for the ensemble classifier.}
\label{tb:exp3_graphrules}
\begin{tabular}{lcc}\toprule
 $r$    & Train TPR, FPR   & Test TPR, FPR  \\ 
\midrule
$x6 > 0.55$ or $x5 \leq -0.09$ and $x6 > -0.88$& $0.97$, $0.003$     & $0.55$, $0.32$ \\
$x6 > -0.88$ and $x2 \leq -0.15$ and $x4 \leq 0.10$   & $0.98$, $0.02$ & $0.68$, $0.27$ \\
$x2 \leq -0.13$ or $x6 > 0.55$ or $x1 \leq -0.07$ & $0.98$, $0.06$     & $0.63$, $0.31$  \\
$x3 \leq 0.02$ and $x4 \leq -0.01$ or $x6 > 0.55$     & $0.98$, $0.03$  & $0.65$, $0.33$ \\
$x1 < 0.04$ and $x6 > 0.88$ and $x5 \leq 0.08$  & $0.91$, $0.01$  & $0.63$, $0.33$\\\bottomrule
\end{tabular}
\end{table*}

Figure~\ref{fig:exp1res_graph} shows the results with graph-based regularization for Example 1. As we can see, compared with Figure~\ref{fig:exp1res}, the accuracy on test and OOD test data are significantly improved. This shows that the decomposition of invariant features from causal graph is sufficient to estimate the target label.
In Table~\ref{tb:exp1_graphrules}, we list basis rules that are used for the classifier. Here, 
$x_1,\ldots,x_5$ are invariant features, and $x_6,\ldots,x_9$ are variant features.
As we can see, the TPR and FPR are relatively worse 
in test environments but still keep good prediction performance. Also because of the using of soft mask for feature selection process, some variant features also contribute to the construction of basis rules. This shows that the soft mask feature selection can regularize the algorithm for highly discriminant patterns as well as keeping stable across different environments.

In Figure~\ref{fig:exp3res_e3_graph}, we plot the results on benchmark `small invariant margin'. With a naive setting of the rule ensemble classifier, the accuracy on training data can reach $1.0$. However, the test and OOD test accuracy are close to a random guess: the model is fully governed by the variant features. Figure~\ref{fig:exp3_e3_graph} shows that after graph regularization, the test and OOD test accuracy are substantially improved. In Table~\ref{tb:exp3_graphrules}, we list several basis rules. These are highly discriminant on training environments, while the TPR and FPR are worse in test environments but still have good prediction performance. Some variants also contribute to the construction of rules. These results show that a soft mask for feature selection based on graph regularization is effective against distributional shift.

\subsection{Experiments on variance-based regularization}

\begin{figure}[t]
 \centering
   \subfloat[env5]{\includegraphics[width=0.46\columnwidth]{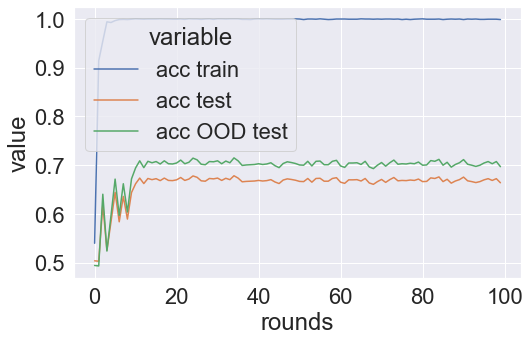}}
   \subfloat[env7]{\includegraphics[width=0.46\columnwidth]{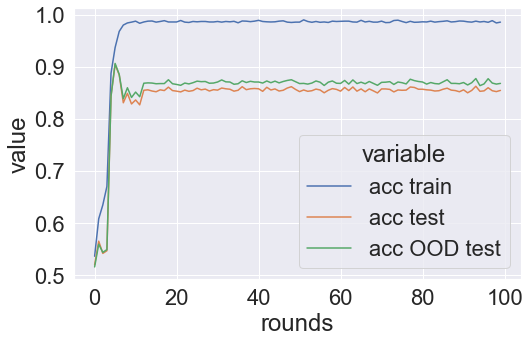}}\\
   \subfloat[env10]{\includegraphics[width=0.46\columnwidth]{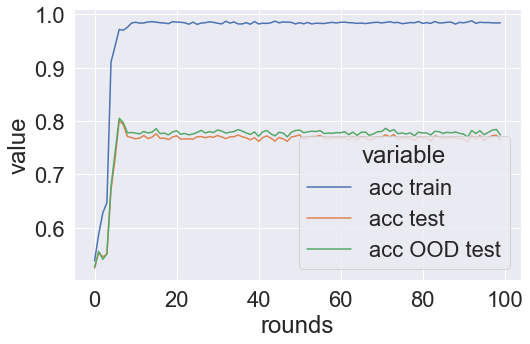}}
   \subfloat[env5]{\includegraphics[width=0.46\columnwidth]{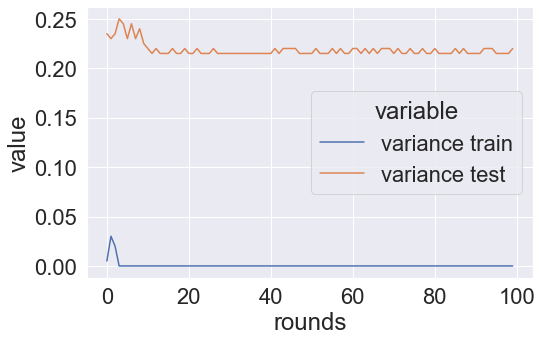}}\\
   \subfloat[env7]{\includegraphics[width=0.46\columnwidth]{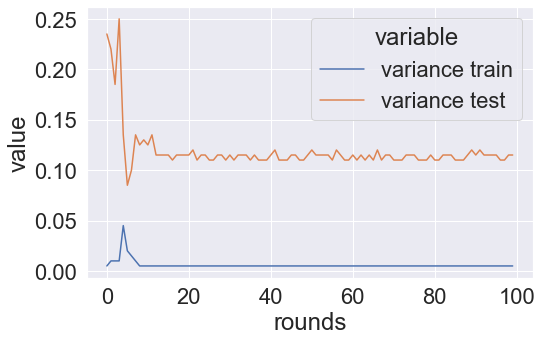}}
   \subfloat[env10]{\includegraphics[width=0.46\columnwidth]{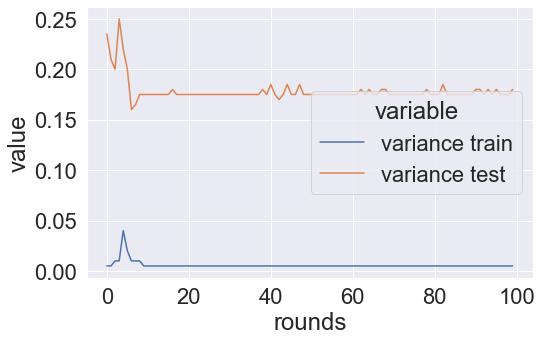}}
    \caption{Performance on `small invariant margin' benchmark. We show that variance-based regularization is effective to improve the robustness.}
    \label{fig:exp3envs}
\end{figure} 

To validate \textbf{RQ2}, we design experiments on one benchmark (`small invariant margin') dataset also used in the previous section, and another benchmark called color MNIST~\cite{arjovsky2019invariant}. For the former benchmark, we construct the artificial feature `ID' according to two groups of invariant features in the original generating process. We test the performance of our method by varying the the number of potential environments $5, 7, 10$, setting the dimensionality of both invariant and variant features to $5$. Figure~\ref{fig:exp3envs} shows that the variance-based regularization is effective in improving the robustness of the model across multiple environments, without knowing the environment labels. Note that the test accuracies on minority data are even a bit better: with the restricting of the regularization term, the rule search algorithm sacrifices accuracy to explore invariant patterns.

\begin{figure}[t]
 \centering
   \subfloat[$\lambda = 1$.\label{fig:cmnist_lmbda_1}]
   {\includegraphics[width=0.46\columnwidth]{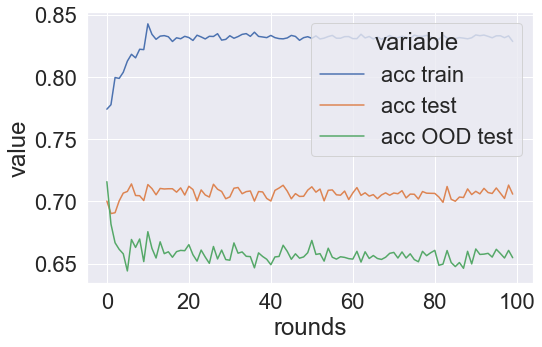}}
   \subfloat[$\lambda = 3$.\label{fig:cmnist_lmbda_3}]
   {\includegraphics[width=0.46\columnwidth]{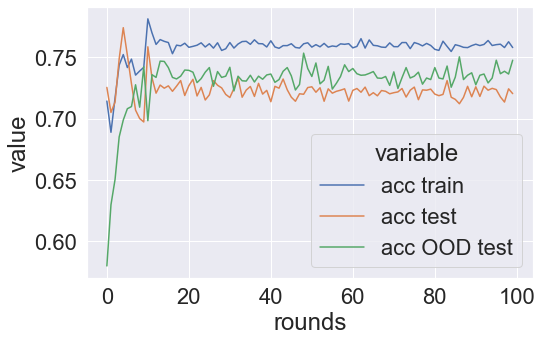}}
   \caption{Accuracy on train, test, OOD test for ColorMNIST benchmark, varying $\lambda$.}
    \label{fig:cmnistlambda}
\end{figure} 

We calculate the variance by constructing additional features regarding to each digit number as same group.
Figure~\ref{fig:cmnistlambda} plots the performance of our method with regularization scale setting to $\lambda=\{1, 3\}$. With lower scale, the algorithm is encouraged to use more discriminant patterns. With larger regularization scale, the variance term restricts the use of variant features, which improves the robustness of the model.

Finally, in order to compare our method with baselines,
we run boosting classifier implemented in sklearn
on those benchmarks (cf.\@ Table~\ref{tb:xgboost}). 

\begin{table}[t]
\centering
\caption{Performance of Traditional Boosting classifier on benchmarks.}
\label{tb:xgboost}
\begin{tabular}{lc}\toprule
 Dataset    &  Traditional Boosting \\ 
\midrule
Example 1    & $0.5235$  \\
Color MNIST       & $0.5004$ \\
`small invariant margin' \phantom{1}3   &  0.5160 \\
`small invariant margin' \phantom{1}5   &  0.5432 \\
`small invariant margin' \phantom{1}7   &  0.3009 \\
`small invariant margin' 10  &  0.5318 \\\bottomrule
\end{tabular} 
\end{table}


\begin{table*}[t]
\centering
\caption{Real-world datasets used in the experiments.}
\label{tb:datasets}
\begin{tabular}{l lrrcccccc}\toprule
\multirow{3}{*}{$i$}&\multirow{3}{*}{$\Omega_i$}&\multirow{3}{*}{$N$}&\multirow{3}{*}{$s$} & \multicolumn{6}{c}{Accuracy}\\
&&&&\multicolumn{2}{c}{Ours}&\multicolumn{2}{c}{GDRO \cite{sagawa2019distributionally}}&\multicolumn{2}{c}{JTT \cite{liu2021just}}\\
&&&& Train & Test  & Train & Test & Train & Test \\ 
\midrule
1 & Adult Income & 32\thinspace561 & 14 & $0.755$ & $0.749$ & $0.798$ & $0.781$ & $0.752$ & $0.714$\\
2 & Bank Marketing & 41\thinspace188 & 20 & $0.867$ & $0.853$ & $0.865$ & $0.850$ & $0.850$ & $0.850$\\
3 & Churn Modeling & 10\thinspace000 & 10 & $0.762$ & $0.749$ & $0.765$ & $0.751$ & $0.782$ & $0.772$ \\
4 & Credit Card & 30\thinspace000 & 23 & $0.719$ & $0.715$ & $0.723$ & $0.690$ & $0.729$ & $0.725$\\
5 & MimicIII Mort & 7\thinspace414 & 25 & $0.818$ & $0.769$ & $0.831$ & $0.789$ & $0.846$ & $0.811$\\
6 & Readmission & 39\thinspace289 & 183 & $0.637$ & $0.621$ & $0.651$ & $0.638$ & $0.655$ & $0.642$\\
\bottomrule
\end{tabular}
\end{table*}

\subsection{Comparison with main baselines on real-world benchmarks}
For \textbf{RQ3}, we compare our algorithms with the latest baselines: GroupDRO~\cite{sagawa2019distributionally} and JTT~\cite{liu2021just}. Table~\ref{tb:datasets} lists results of this comparison on real-world datasets. Contrary to the original work of the baselines, which work for the overparameterized neural networks, here we re-implement those baselines for the boosting ensemble algorithms, for which even weak rules can construct strong classifiers. The main advantages of a rule ensemble is that the discovered decision rules are both discriminant and interpretable. Hence, domain expert can easily make use of the discovered rules to understand the behavior of the classifiers. For instance, in the MimicIII dataset, the top rules that are used to form the ensemble by our algorithm are: `PTT $> 40.5$ and LACTATE $> 2.21$' with TPR $= 0.49$ and FPR $= 0.08$, `BUN $> 18.3$ and Marital Status Cat $\leq 3.0$' with TPR $= 0.47$ and FPR $= 0.06$. On the other hand, GroupDRO extracted rules `BUN $> 35.6$ and ALBUMIN $\leq 2.6$' with TPR $= 0.25$ and FRP $= 0.08$, `ALBUMIN $\leq 2.28$' with TPR $= 0.23$ and FPR $= 0.09$. JTT extracts rules `ANIONGAP $> 17.13$ and PHOSPHATE $\leq 2.52$' with TPR $= 0.55$ and FPR $= 0.16$. We can see that different baselines discovered different rules but both achieved high accuracy, which shows that our algorithm has competitive performance. 

\section{Related Work}
Previous work on spurious correlation mainly focuses on uncovering the distributional shifts across different groups when group representations change between train and test environments~\cite{sinha2017certifying,ben2013robust}. Some work leverage distributional robust optimization (DRO) methods which define the uncertainty set as a divergence ball over the training distribution to ensure the robustness~\cite{duchi2021statistics,miyato2018virtual}. Some other work employ a regularizer to restrict the targeted loss to improve the robustness~\cite{shafieezadeh2015distributionally}. Group DRO considers the group information for the worst case performance of the model, considering the label shifts~\cite{hu2018does}, data shifts~\cite{oren2019distributionally}, or overparameterized scenarios with vanishing training loss and worst group performance case~\cite{sagawa2019distributionally}. All these methods require the annotation of group information to perform the optimization process. On this other hand, some settings do not provide group information during training steps. Only few samples on the validation data are allowed to have the group information. Methods like Just Train Twice (JTT)~\cite{liu2021just} employ a two step training strategy to highlight the worst case samples to improve the robustness. Other methods focus on learning group information by first clustering the samples to improve the robustness~\cite{sohoni2020no}, or proposing an auditor with a pre-defined complex class to search high-loss groups to minimize the identified discrepancies~\cite{kim2019multiaccuracy}.

A causal graph is a directed acyclic graph (DAG) that represents causes and effects relationships which can be specified as Structural Causal Models (SCMs). SCMs represent certain distributions that govern the generating process of the data. An intervention is introduced by the form of modification of the mechanism on one or more variables in the causal graph. Such changes indicate the interventional shifts for which the causal invariance relation can be derived. Some work makes use of this invariance to invent the causal invariant predictor (ICP)~\cite{peters2016causal}. Invariant Risk Minimization (IRM) makes use of a weaker form of causal invariance to learn an invariant encoder to make sure the downstream classifier is optimal across different environments~\cite{arjovsky2019invariant}. Aiming at Out-of-distribution (OOD) generalization, Risk Extrapolation (REx) proposes a variance penalty method for model agnostic and deep neural nets regarding to the invariance of risks~\cite{krueger2021out}.

Ensemble learning is regarded as the most powerful method for predictive modeling. Ensemble predictions is constructed of a linear combination of base ensemble learners~\cite{friedman2008predictive}. Some methods focuses on the generation of decision rules to achieve highly discriminative performance~\cite{belfodil2018anytime}. Boosting methods are also adopted to construct the ensembles with rule based learners~\cite{schapire2013explaining}. Gradient Boosting Decision Trees (GBDT) plays an important role for ensemble models, with popular implementations like XGBoost~\cite{chen2016xgboost}. Some implementations focus on generating the decision trees with fewer data points and features, that improve the training efficiency of the ensemble model~\cite{ke2017lightgbm}. However, these model do not tackle the distributional shifts in the datasets hence achieves worst case performance across different environments. Our model considers the environmental shifts and provides a new way to construct predictive and robust rules.

\section{Discussion and Conclusion}


We study the problem of robustness of local decision rule ensembles. Local decision rules could provide both high predictive performance and explainability. However, the stability of those explanations given by the rules is still underexplored. Our work aims to fill this gap. We start from analyzing the robustness of decision rule ensembles guided by potential causal graph.
Empirical results show how the performance of traditional rule ensembles vary across environments. The distributional shifts are assumed to be caused by potential interventions on the underlying system. We further propose two regularization terms to improve the robustness of decision rule ensembles. The graph-based term is built by decomposing invariant features using a given causal graph; the variance-based term relies on an additional artificial feature that can restrict the model's decision boundary within groups. We conduct experiments on several synthetic and benchmark datasets. The experimental results show that our method can significantly improve the robustness of rule ensembles across environments.


\begin{thebibliography}{10}

\bibitem{morik2005local}
K.~Morik and J.-F. Boulicaut, {\em Local Pattern Detection: International
  Seminar Dagstuhl Castle, Germany, April 12-16, 2004, Revised Selected
  Papers}, vol.~3539.
\newblock Springer Science \& Business Media, 2005.

\bibitem{furnkranz2012foundations}
J.~F{\"u}rnkranz, D.~Gamberger, and N.~Lavra{\v{c}}, {\em Foundations of rule
  learning}.
\newblock Springer Science \& Business Media, 2012.

\bibitem{belfodil2018anytime}
A.~Belfodil, A.~Belfodil, and M.~Kaytoue, ``Anytime subgroup discovery in
  numerical domains with guarantees,'' in {\em ECML-PKDD}, pp.~500--516,
  Springer, 2019.

\bibitem{chen2016xgboost}
T.~Chen and C.~Guestrin, ``Xgboost: A scalable tree boosting system,'' in {\em
  SIGKDD}, pp.~785--794, 2016.

\bibitem{kalofolias2017efficiently}
J.~Kalofolias, M.~Boley, and J.~Vreeken, ``Efficiently discovering locally
  exceptional yet globally representative subgroups,'' in {\em ICDM},
  pp.~197--206, 2017.

\bibitem{du2020fairness}
X.~Du, Y.~Pei, W.~Duivesteijn, and M.~Pechenizkiy, ``Fairness in network
  representation by latent structural heterogeneity in observational data,'' in
  {\em AAAI}, vol.~34, pp.~3809--3816, 2020.

\bibitem{budhathoki2021discovering}
K.~Budhathoki, M.~Boley, and J.~Vreeken, ``Discovering reliable causal rules,''
  in {\em SDM}, pp.~1--9, SIAM, 2021.

\bibitem{fischer2021s}
J.~Fischer, A.~Olah, and J.~Vreeken, ``What’s in the box? exploring the inner
  life of neural networks with robust rules,'' in {\em ICML}, pp.~3352--3362,
  PMLR, 2021.

\bibitem{zeng2021causal}
S.~Zeng, M.~A. Bayir, J.~J. Pfeiffer~III, D.~Charles, and E.~Kiciman, ``Causal
  transfer random forest: Combining logged data and randomized experiments for
  robust prediction,'' in {\em WSDM}, pp.~211--219, 2021.

\bibitem{devos2021versatile}
L.~Devos, W.~Meert, and J.~Davis, ``Versatile verification of tree ensembles,''
  in {\em ICML}, pp.~2654--2664, PMLR, 2021.

\bibitem{kull2014patterns}
M.~Kull and P.~Flach, ``Patterns of dataset shift,'' in {\em First
  International Workshop on Learning over Multiple Contexts (LMCE) at
  ECML-PKDD}, vol.~5, 2014.

\bibitem{duchi2021learning}
J.~C. Duchi and H.~Namkoong, ``Learning models with uniform performance via
  distributionally robust optimization,'' {\em The Annals of Statistics},
  vol.~49, no.~3, pp.~1378--1406, 2021.

\bibitem{gong2016domain}
M.~Gong, K.~Zhang, T.~Liu, D.~Tao, C.~Glymour, and B.~Sch{\"o}lkopf, ``Domain
  adaptation with conditional transferable components,'' in {\em ICML},
  pp.~2839--2848, PMLR, 2016.

\bibitem{magliacane2017domain}
S.~Magliacane, T.~van Ommen, T.~Claassen, S.~Bongers, P.~Versteeg, and J.~M.
  Mooij, ``Domain adaptation by using causal inference to predict invariant
  conditional distributions,'' {\em arXiv preprint arXiv:1707.06422}, 2017.

\bibitem{arjovsky2019invariant}
M.~Arjovsky, L.~Bottou, I.~Gulrajani, and D.~Lopez-Paz, ``Invariant risk
  minimization,'' {\em arXiv preprint arXiv:1907.02893}, 2019.

\bibitem{peters2017elements}
J.~Peters, D.~Janzing, and B.~Sch{\"o}lkopf, {\em Elements of causal inference:
  foundations and learning algorithms}.
\newblock The MIT Press, 2017.

\bibitem{friedman2008predictive}
J.~H. Friedman and B.~E. Popescu, ``Predictive learning via rule ensembles,''
  {\em The Annals of Applied Statistics}, vol.~2, no.~3, pp.~916--954, 2008.

\bibitem{duivesteijn2016exceptional}
W.~Duivesteijn, A.~J. Feelders, and A.~Knobbe, ``Exceptional model mining,''
  {\em Data Mining and Knowledge Discovery}, vol.~30, no.~1, pp.~47--98, 2016.

\bibitem{furnkranz2005roc}
J.~F{\"u}rnkranz and P.~A. Flach, ``Roc ‘n’rule learning—towards a better
  understanding of covering algorithms,'' {\em Machine learning}, vol.~58,
  no.~1, pp.~39--77, 2005.

\bibitem{chandrashekar2014survey}
G.~Chandrashekar and F.~Sahin, ``A survey on feature selection methods,'' {\em
  Computers \& Electrical Engineering}, vol.~40, no.~1, pp.~16--28, 2014.

\bibitem{correa2019statistical}
J.~D. Correa and E.~Bareinboim, ``From statistical transportability to
  estimating the effect of stochastic interventions,'' in {\em IJCAI},
  pp.~1661--1667, 2019.

\bibitem{pearl2009causality}
J.~Pearl, {\em Causality}.
\newblock Cambridge university press, 2009.

\bibitem{tian2002testable}
J.~Tian and J.~Pearl, ``On the testable implications of causal models with
  hidden variables,'' in {\em UAI}, pp.~519--527, 2002.

\bibitem{maddison2016concrete}
C.~J. Maddison, A.~Mnih, and Y.~W. Teh, ``The concrete distribution: A
  continuous relaxation of discrete random variables,'' {\em arXiv preprint
  arXiv:1611.00712}, 2016.

\bibitem{yamada2020feature}
Y.~Yamada, O.~Lindenbaum, S.~Negahban, and Y.~Kluger, ``Feature selection using
  stochastic gates,'' in {\em ICML}, pp.~10648--10659, PMLR, 2020.

\bibitem{cestnik1990estimating}
B.~Cestnik, ``Estimating probabilities: A crucial task in machine learning,''
  in {\em ECAI}, pp.~147--149, 1990.

\bibitem{meeng2021for}
M.~Meeng and A.~J. Knobbe, ``For real: a thorough look at numeric attributes in
  subgroup discovery,'' {\em DMKD}, vol.~35, no.~1, pp.~158--212, 2021.

\bibitem{janzing2019causal}
D.~Janzing, ``Causal regularization,'' {\em arXiv preprint arXiv:1906.12179},
  2019.

\bibitem{heinze2021conditional}
C.~Heinze-Deml and N.~Meinshausen, ``Conditional variance penalties and domain
  shift robustness,'' {\em Machine Learning}, vol.~110, no.~2, pp.~303--348,
  2021.

\bibitem{aubin2021linear}
B.~Aubin, A.~S{\l}owik, M.~Arjovsky, L.~Bottou, and D.~Lopez-Paz, ``Linear
  unit-tests for invariance discovery,'' {\em arXiv preprint arXiv:2102.10867},
  2021.

\bibitem{parascandolo2020learning}
G.~Parascandolo, A.~Neitz, A.~Orvieto, L.~Gresele, and B.~Sch{\"o}lkopf,
  ``Learning explanations that are hard to vary,'' {\em arXiv preprint
  arXiv:2009.00329}, 2020.

\bibitem{beery2018recognition}
S.~Beery, G.~Van~Horn, and P.~Perona, ``Recognition in terra incognita,'' in
  {\em Proceedings of the European conference on computer vision (ECCV)},
  pp.~456--473, 2018.

\bibitem{wall2003singular}
M.~E. Wall, A.~Rechtsteiner, and L.~M. Rocha, ``Singular value decomposition
  and principal component analysis,'' in {\em A practical approach to
  microarray data analysis}, pp.~91--109, Springer, 2003.

\bibitem{nastl2024causal}
V.~Y. Nastl and M.~Hardt, ``Do causal predictors generalize better to new
  domains?,'' in {\em The Thirty-eighth Annual Conference on Neural Information
  Processing Systems}, 2024.

\bibitem{adult_2}
B.~Becker and R.~Kohavi, ``{Adult}.'' UCI Machine Learning Repository, 1996.
\newblock {DOI}: https://doi.org/10.24432/C5XW20.

\bibitem{bank_marketing_222}
S.~Moro, P.~Rita, and P.~Cortez, ``{Bank Marketing}.'' UCI Machine Learning
  Repository, 2014.
\newblock {DOI}: https://doi.org/10.24432/C5K306.

\bibitem{default_of_credit_card_clients_350}
I.-C. Yeh, ``{Default of Credit Card Clients}.'' UCI Machine Learning
  Repository, 2009.
\newblock {DOI}: https://doi.org/10.24432/C55S3H.

\bibitem{johnson2016mimic}
A.~E. Johnson, T.~J. Pollard, L.~Shen, L.-w.~H. Lehman, M.~Feng, M.~Ghassemi,
  B.~Moody, P.~Szolovits, L.~Anthony~Celi, and R.~G. Mark, ``Mimic-iii, a
  freely accessible critical care database,'' {\em Scientific data}, vol.~3,
  no.~1, pp.~1--9, 2016.

\bibitem{gardner2024benchmarking}
J.~Gardner, Z.~Popovic, and L.~Schmidt, ``Benchmarking distribution shift in
  tabular data with tableshift,'' {\em Advances in Neural Information
  Processing Systems}, vol.~36, 2024.

\bibitem{sagawa2019distributionally}
S.~Sagawa, P.~W. Koh, T.~B. Hashimoto, and P.~Liang, ``Distributionally robust
  neural networks for group shifts: On the importance of regularization for
  worst-case generalization,'' {\em arXiv preprint arXiv:1911.08731}, 2019.

\bibitem{liu2021just}
E.~Z. Liu, B.~Haghgoo, A.~S. Chen, A.~Raghunathan, P.~W. Koh, S.~Sagawa,
  P.~Liang, and C.~Finn, ``Just train twice: Improving group robustness without
  training group information,'' in {\em ICML}, pp.~6781--6792, PMLR, 2021.

\bibitem{sinha2017certifying}
A.~Sinha, H.~Namkoong, R.~Volpi, and J.~Duchi, ``Certifying some distributional
  robustness with principled adversarial training,'' {\em arXiv preprint
  arXiv:1710.10571}, 2017.

\bibitem{ben2013robust}
A.~Ben-Tal, D.~Den~Hertog, A.~De~Waegenaere, B.~Melenberg, and G.~Rennen,
  ``Robust solutions of optimization problems affected by uncertain
  probabilities,'' {\em Management Science}, vol.~59, no.~2, pp.~341--357,
  2013.

\bibitem{duchi2021statistics}
J.~C. Duchi, P.~W. Glynn, and H.~Namkoong, ``Statistics of robust optimization:
  A generalized empirical likelihood approach,'' {\em Mathematics of Operations
  Research}, vol.~46, no.~3, pp.~946--969, 2021.

\bibitem{miyato2018virtual}
T.~Miyato, S.-i. Maeda, M.~Koyama, and S.~Ishii, ``Virtual adversarial
  training: a regularization method for supervised and semi-supervised
  learning,'' {\em IEEE transactions on pattern analysis and machine
  intelligence}, vol.~41, no.~8, pp.~1979--1993, 2018.

\bibitem{shafieezadeh2015distributionally}
S.~Shafieezadeh~Abadeh, P.~M. Mohajerin~Esfahani, and D.~Kuhn,
  ``Distributionally robust logistic regression,'' {\em Advances in neural
  information processing systems}, vol.~28, 2015.

\bibitem{hu2018does}
W.~Hu, G.~Niu, I.~Sato, and M.~Sugiyama, ``Does distributionally robust
  supervised learning give robust classifiers?,'' in {\em International
  Conference on Machine Learning}, pp.~2029--2037, PMLR, 2018.

\bibitem{oren2019distributionally}
Y.~Oren, S.~Sagawa, T.~B. Hashimoto, and P.~Liang, ``Distributionally robust
  language modeling,'' {\em arXiv preprint arXiv:1909.02060}, 2019.

\bibitem{sohoni2020no}
N.~Sohoni, J.~Dunnmon, G.~Angus, A.~Gu, and C.~R{\'e}, ``No subclass left
  behind: Fine-grained robustness in coarse-grained classification problems,''
  {\em Advances in Neural Information Processing Systems}, vol.~33,
  pp.~19339--19352, 2020.

\bibitem{kim2019multiaccuracy}
M.~P. Kim, A.~Ghorbani, and J.~Zou, ``Multiaccuracy: Black-box post-processing
  for fairness in classification,'' in {\em Proceedings of the 2019 AAAI/ACM
  Conference on AI, Ethics, and Society}, pp.~247--254, 2019.

\bibitem{peters2016causal}
J.~Peters, P.~B{\"u}hlmann, and N.~Meinshausen, ``Causal inference by using
  invariant prediction: identification and confidence intervals,'' {\em Journal
  of the Royal Statistical Society Series B: Statistical Methodology}, vol.~78,
  no.~5, pp.~947--1012, 2016.

\bibitem{krueger2021out}
D.~Krueger, E.~Caballero, J.-H. Jacobsen, A.~Zhang, J.~Binas, D.~Zhang,
  R.~Le~Priol, and A.~Courville, ``Out-of-distribution generalization via risk
  extrapolation (rex),'' in {\em International conference on machine learning},
  pp.~5815--5826, PMLR, 2021.

\bibitem{schapire2013explaining}
R.~E. Schapire, ``Explaining adaboost,'' in {\em Empirical inference:
  festschrift in honor of vladimir N. Vapnik}, pp.~37--52, Springer, 2013.

\bibitem{ke2017lightgbm}
G.~Ke, Q.~Meng, T.~Finley, T.~Wang, W.~Chen, W.~Ma, Q.~Ye, and T.-Y. Liu,
  ``Lightgbm: A highly efficient gradient boosting decision tree,'' {\em
  NeurIPS}, vol.~30, pp.~3146--3154, 2017.

\end{thebibliography}

\appendices

\section{Invariant feature decomposition}\label{app:ifd}

The core idea is that
we can decompose the joint distribution into product conditional distributions of each c-component~\cite{correa2019statistical,tian2002testable,pearl2009causality}.
Each variable in c-component is only dependent on its non-descendant
variables in the c-component and the effective parents of
its non-descendant variables in the c-component.
We derive the following theorem for causal invariant decomposition:

\begin{theorem}\label{tm:cid}
Let $X, Y \subseteq V$ be disjoint sets of variables. Let $W = An(X \cup Y)_{\mathcal{G}}$ be partitioned into c-components $C(W) = \{C_1(W_1),\ldots,C_H(W_H)\}$ in causal graph $\mathcal{G}_{[An(X \cup Y)]}$.
Let $A$ be a set of manipulable variables for potential interventions. Then $X^{\prime} \in C_h(W_h)$ is an invariant feature across interventional environments when either of the following conditions holds:
\begin{equation}
C_h(W_h) \cap A = \emptyset,
\end{equation}
\begin{equation}
C_h(W_h) \cap De(Y) = \emptyset,
\end{equation}
where $De(Y)$ represents the descendants of $Y$.
\end{theorem}

\begin{proof}
First, from Assumption~1 we can know $A \cap Pa(Y) = \emptyset$. 
When $C_h(W_h) \cap A = \emptyset$, there is no manipulable variable in $C_h(W_h$. Because each variable in c-component is only dependent on its non-descendant variables in the c-component and the effective parents of its non-descendant variables in the c-component, we have $P(Y|X_h, A) = P(Y|X_h)$. When $C_h(W_h) \cap De(Y) = \emptyset$, $X_h$ can block the path from $A_h$ to $Y$, then we have $P(Y|X_h,A_h) = P(Y|X_h)$.
\end{proof} 

\begin{algorithm}[t]
\caption{Invariant feature decomposition.}
\label{al:alg2}
\begin{algorithmic}[1]
\Require Dataset $\Omega$, Graph $\mathcal{G}$, c-components decomposition operator $\psi$
\Ensure $X^{\prime}$
\Function{CCS}{$\Omega, \mathcal{G}$, $\psi$}
  \State $X^{\prime} \leftarrow \{\}$;
  \State $W \leftarrow An(X \cup Y)_{\mathcal{G}}$;
  \State $C_1(W_1), \cdots, C_H(W_H) \leftarrow \psi(G_{[An(X \cup Y)]})$;
  \For{$h=1$ to $H$}
    \If{$C_h(W_h) \cap A = \emptyset$}
    \State $X^{\prime} \leftarrow C_h(W_h)$;
    \EndIf
    \If{$C_h(W_h) \cap De(Y) = \emptyset$}
    \State $X^{\prime} \leftarrow C_h(W_h) \setminus A$
    \EndIf
  \EndFor
  \State \Return{$X^{\prime}$};
\EndFunction
\end{algorithmic}
\end{algorithm}

We developed Algorithm~\ref{al:alg2} to find set of invariant features $X^{\prime}$. The c-component decomposition operator $\psi$ is based on~\cite{tian2002testable}.

\section{Variance-based regularization}\label{app:vbr}
For each single decision rule, we have $r(x) \in \{-1,1\}$. Hence we have $F(x) = \sum_{m=1}^M \alpha_m r_m(x) \in \mathbb{R}$. Then we can map the output to $[0,1]$ by applying the sigmoid funciton. Here we abuse $F(x)$ as $Sigmoid(F(x))$. Here we recall:

\begin{equation*}\label{eq:cvarf}
C_{F, \nu} = E[\text{Var}(F(X)|Y, ID)^{\nu}], 
\end{equation*}

specifically, following~\cite{heinze2021conditional}, we can compute $C_{F,1}$ as:
\begin{equation*}
\hat{C}_{F,1} = \frac{1}{q}\sum_{j=1}^q \frac{1}{|G_j|} \sum_{i\in G_j}(F(x_i) - \hat{\mu}_{F,j})^2,
\end{equation*}
where
\begin{equation*}
\hat{\mu}_{F,j} = \frac{1}{|G_j|} \sum_{i \in G_j} F(x_i).
\end{equation*}
This variance term indicates that the decision boundary of $F(x)$ for data points in same group by artificial feature and label should be similar, quantifying by the predictive probability.

\end{document}